\documentclass{hld2023}
\usepackage[utf8]{inputenc}
\usepackage{setspace} 
\usepackage{amsfonts} 
\usepackage{bm}
\usepackage{lineno} 
\usepackage{float}
\usepackage{caption}
\usepackage{pdflscape}
\usepackage{cleveref}
\usepackage{ifthen}
\usepackage{breqn}
\usepackage[hide=false,setmargin=true,marginparwidth=0.7in]{marginalia}
\usepackage{multirow}
\usepackage{multicol}

\LinesNumbered

\newcommand{\ph}{{\varphi}}

\DeclareMathOperator{\var}{\mathbf{Var}}

\newtheorem{approximation}{Approximation}

\title[Network Degeneracy as an Indicator of Training Performance]{Network Degeneracy as an Indicator of Training Performance: Comparing Finite and Infinite Width Angle Predictions}

\hldauthor{
\Name{Cameron Jakub} \Email{cjakub@uoguelph.ca}\\
\addr {University of Guelph, Ontario, Canada} 
\AND
\Name{Mihai Nica} \Email{nicam@uoguelph.ca}\\
\addr {University of Guelph, Ontario, Canada} }

\begin{document}

\maketitle

\begin{abstract}

Neural networks are powerful functions with widespread use, but the theoretical behaviour of these functions is not fully understood. Creating \emph{deep} neural networks by stacking many layers has achieved exceptional performance in many applications and contributed to the recent explosion of these methods. Previous works have shown that depth can exponentially increase the expressibility of the network \cite{poole_expressivity, eldan_expressivity}. However, as networks get deeper and deeper, they are more susceptible to becoming \emph{degenerate}. We observe this degeneracy in the sense that on initialization, inputs tend to become more and more correlated as they travel through the layers of the network. If a network has too many layers, it tends to approximate a (random) constant function, making it effectively incapable of distinguishing between inputs. This seems to affect the training of the network and cause it to perform poorly, as we empirically investigate in this paper.
We use a simple algorithm that can accurately predict the level of degeneracy for any given fully connected ReLU network architecture, and demonstrate how the predicted degeneracy relates to training dynamics of the network. We also compare this prediction to predictions derived using infinite width networks. 
\end{abstract}

\section{Introduction and Main Results}
\label{sec:main_results}

Our previous work \emph{Depth Degeneracy in Neural Networks: Vanishing Angles in Fully Connected ReLU Networks} \cite{jakub_nica} theoretically studied the ``large depth degeneracy" phenomenon for finite width ReLU networks.
This workshop paper extends the work of that paper, and uses our previous theoretical results as an input into experiments that investigate how the level of degeneracy can influence training. Consider two inputs fed into an initialized feed-forward ReLU network with depth $L$ and layer widths $n_\ell,\; 1 \leq \ell \leq L$ (see \Cref{app:network_definition} for a full definition of the network). We assume the network is initialized with independent Gaussian weights  so that the network is on the ``edge of chaos'' \cite{hayou, schoenholz}, and that the angle between inputs is defined using the inner product on $\mathbb{R}^{n_\ell}$ in the standard way. Given this setup, \Cref{algo:update_rule} (established theoretically in \cite{jakub_nica}) provides us with a simple method to accurately predict the angle between those inputs after travelling through the layers of the network on network initialization up to an error of size $\mathcal{O}(n_\ell^{-2})$ in layer $\ell$.


\begin{algorithm2e}
\caption{Angle prediction between inputs for a feed-forward ReLU network with depth $L$ and layer widths $n_\ell,\; 1 \leq \ell \leq L$. The function $\mu(\theta,n)$ is given in Theorem \ref{thm:mean_var_exp}.}
\label{algo:update_rule}
$\theta^0 = $ angle between inputs \;
\For{$\ell = 0, \ldots, L-1$}{
$x = \mu(\theta^\ell, n_\ell)$ \tcp*{$x$ represents 
 $\mathbf{E}[\ln(\sin^2(\theta^{\ell+1}))]$}
$\theta^{\ell+1} = \arcsin(e^{\frac{x}{2}})$ \;} 
Final angle $= \theta^L$
\end{algorithm2e}

\begin{figure}[H]
    \centering
    \includegraphics[scale=0.55]{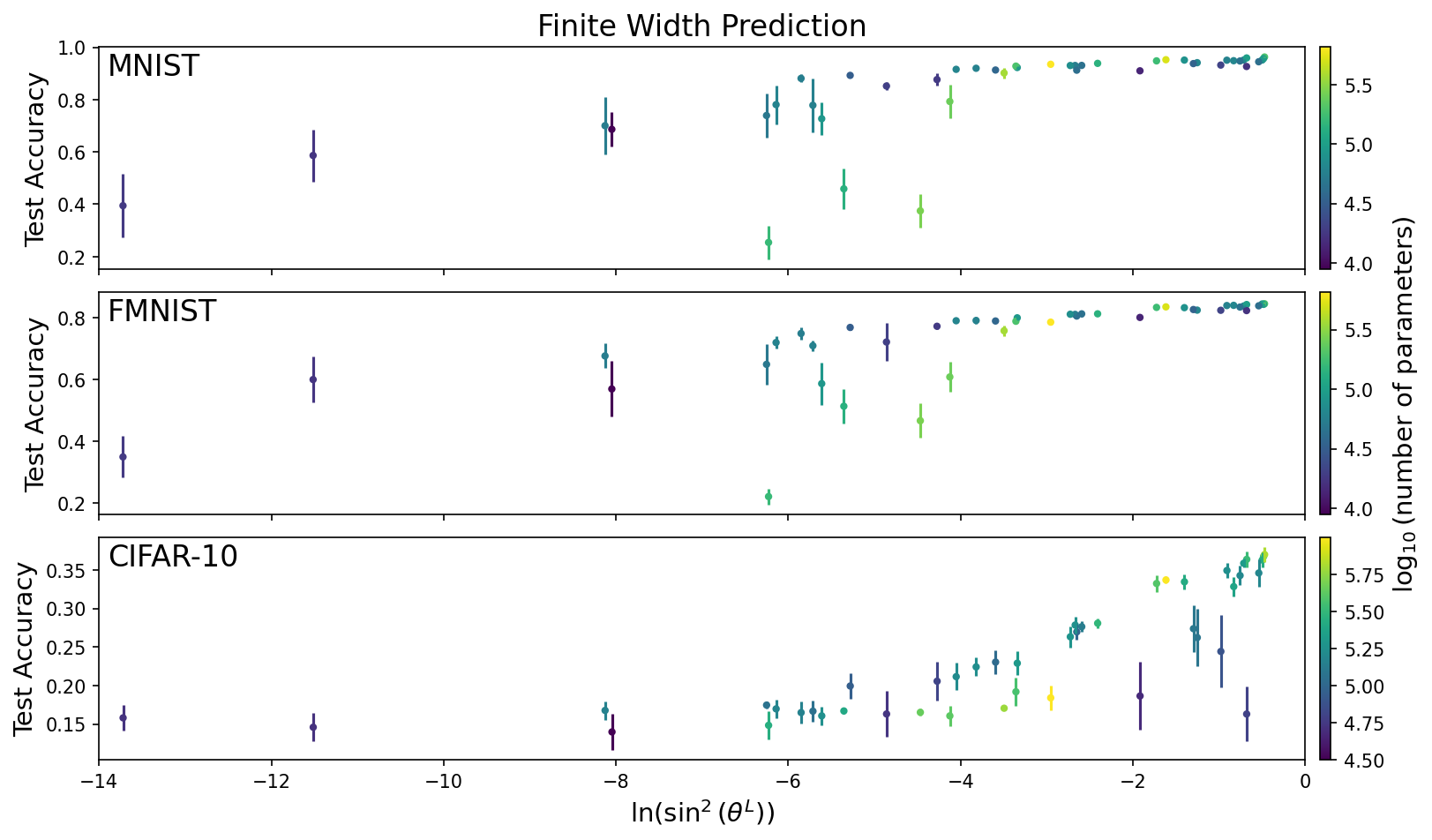}
    \caption{We compare 45 different network architectures trained on the MNIST \cite{mnist}, Fashion-MNIST \cite{fmnist}, and CIFAR-10 \cite{cifar} datasets 10 times each. Using the architecture of the network and \Cref{algo:update_rule}, we predict the angle between 2 orthogonal inputs at the final output layer of the network on initialization. We express the angle as $\ln(\sin^2(\theta^L))$, to follow the form used when  developing the finite width approximations. The angle is plotted against the accuracy of each network on the test data after training, with error bars representing a 95\% confidence interval across the 10 runs. All networks are trained using 1 epoch, batch size $=100$, categorical cross-entropy loss, the ADAM optimizer, and default learning rate in the Keras module of TensorFlow \cite{tensorflow}. See \Cref{app:network_architectures} for details on all of the network architectures used.}
    \label{fig:simulations}
\end{figure}

\Cref{algo:update_rule} predicts the angle at the final layer on initialization based solely on the network architecture $n_1,n_2,\ldots n_L$. If all inputs into an initialized network tend to be highly correlated by the final layer, this could make it difficult for the network to distinguish the differences between inputs and therefore harder to train. \Cref{fig:simulations} demonstrates how networks which exhibit this type of degeneracy empirically tend to perform worse \emph{after training}, and seem to train less consistently than networks which can better distinguish between inputs on initialization.

When \Cref{algo:update_rule} predicts that the network architecture forces inputs to become highly correlated on initialization, this serves a warning that the network may train poorly. Before going through the computationally expensive process of training many networks to assess their performance, this prediction could be used to quickly filter out network architectures that are unlikely to perform well. The simplicity and efficiency of the algorithm may lend itself well to applications in neural architecture search, and would be an interesting starting point for more detailed experiments and/or theoretical explanations about training.


\subsection{Finite Width - Small Angle Evolution}
Since the effect of each layer is independent of everything previous, $\theta^\ell$ can be thought of as a Markov chain evolving as layer number $\ell$ increases. As expected by the aforementioned ``large depth degeneracy'' phenomenon, the angle $\theta^\ell$ tends towards $0$ as the current layer $\ell$ goes towards infinity. This indicates that the hidden layer representation of \emph{any} two inputs in a deep neural network becomes closer and closer to co-linear as depth increases. We found a simple update rule in \citet{jakub_nica} which predicts how the angle between inputs evolves, given below in \Cref{approx:simple}. The algorithm works well for finite sized networks because the errors are controlled up to size $\mathcal{O}(n^{-2})$ in the layer sizes.

\begin{approximation}{(Finite width small angle update rule)}
\label{approx:simple}
For small angles $\theta^\ell \ll 1$ and large layer widths $n_\ell \gg 1$, the angle $\theta^{\ell+1}$ at layer $\ell+1$ is well approximated by
\begin{equation} \label{eq:update_simple}
\ln \sin^2(\theta^{\ell+1}) \approx
\ln \sin^2(\theta^\ell) - \frac{2}{3\pi}\theta^\ell - \rho(n_\ell), 
\end{equation} 
where $\rho(n_\ell)$ is a constant which depends on the width $n_\ell$ of layer $\ell$, namely:
\begin{equation}
    \rho(n) := \ln\left( \frac{n+5}{n-1}\right) - \frac{10n}{\left(n+5\right)^2} + \frac{6n}{\left(n-1\right)^2 }= \frac{2}{n} + \mathcal{O}\left(n^{-2}\right). \label{eq:rho}
\end{equation}
\end{approximation}

\begin{remark}
    Approximation \ref{approx:simple}  comes from a simplification of more precise formulas for the mean and variance of the variable $\ln(\sin^2(\theta^\ell))$, which are stated in  \Cref{thm:mean_var_exp}. Specifically, \Cref{approx:simple} is derived from a linear approximation of $\mu(\theta,n)$ in \Cref{thm:mean_var_exp} about $\theta=0$. For $\theta^\ell$ sufficiently small, line 3 of \Cref{algo:update_rule} could be replaced with the simpler linear update rule given in \Cref{approx:simple}: $x = \ln \sin^2 \theta^\ell - \frac{2}{3\pi}\theta^\ell - \rho(n_\ell)$.
\end{remark}

\subsection{Comparison to Infinite Width Networks}
The angle degeneracy phenomenon has been studied in previous works for networks in the limit of infinite width \cite{hayou, schoenholz, principles_deep_learning, boris_correlation_functions}. The infinite width case uses the law of large numbers and thereby disregards any random fluctuations in $\theta^{\ell+1}$ given $\theta^\ell$. These random fluctuations, though small, can accumulate over many layers leading to inaccurate predictions for finite width networks (see \Cref{updaterule}). The infinite width update rule is given below in \Cref{approx:infinite}.

\begin{approximation}{(Infinite width update rule)}
\label{approx:infinite}
In the limit that the width of each layer tends to infinity, the \emph{infinite width approximation} for the angle $\theta^{\ell+1}$ given $\theta^\ell$ is    
\begin{equation} \label{eq:infinte_width_update}
\cos\left( \theta^{\ell+1} \right) =\frac{\sin(\theta^\ell)+(\pi-\theta^\ell)\cos(\theta^\ell)}{\pi}. 
\end{equation}
\end{approximation}

\begin{figure}[H]
    \centering
    \includegraphics[scale=0.48]{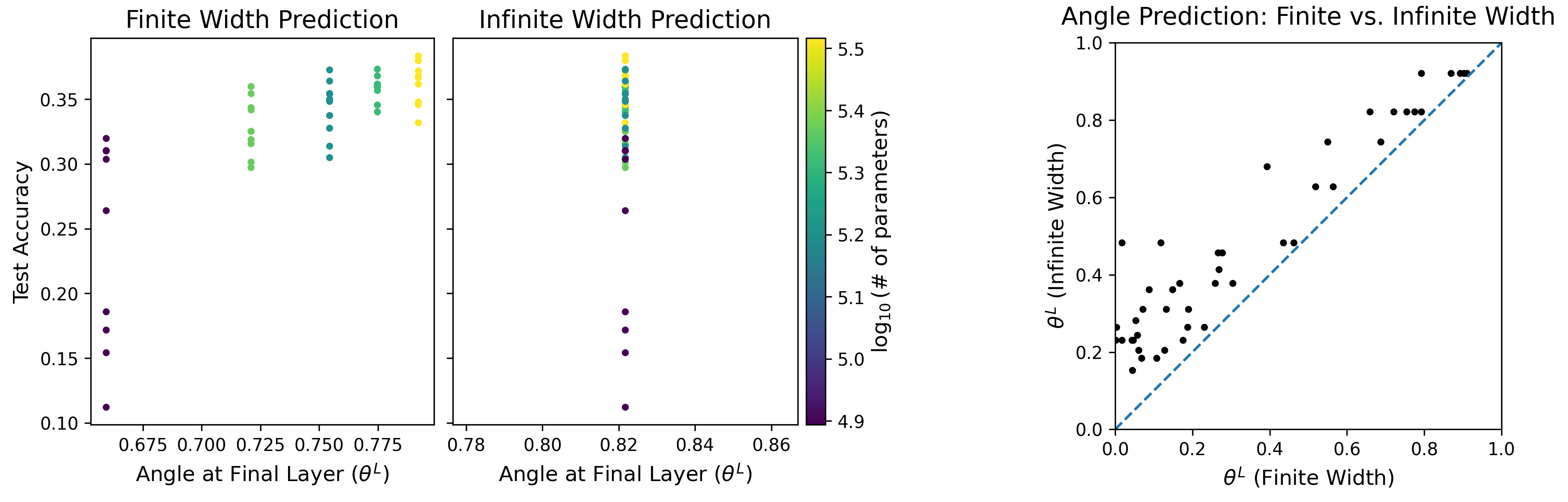}
    \caption{Left: Comparison of the finite and infinite width predictions for 5 network architectures with a depth of $L=3$ trained 10 times each on the CIFAR-10  dataset \cite{cifar}. The infinite width predicts the same final angle for all networks, since it only depends on network depth. Right: Using the same 45 network architectures as in \Cref{fig:simulations}, we plot a comparison of the predicted angle $\theta^L$ using \Cref{algo:update_rule} (finite width) versus the infinite width prediction. We see that the infinite width prediction tends to underestimate the rate at which $\theta^\ell$ tends towards 0.}
    \label{fig:comparison}
\end{figure}

Another issue with using the infinite width prediction to study finite width networks is that all networks with the same depth are treated exactly the same, since it does not take into account the width of each layer. Both the depth of the network and the width of each layer affect how the angle between inputs propagates layer-by-layer through the network. \Cref{fig:comparison}-Left illustrates how our method yields different angle predictions for different architectures with the same depth, while the infinite width method does not. \Cref{fig:comparison}-Right shows the how the infinite width predictions differ from our ``finite width'' method which takes into account fluctuations of size $\mathcal{O}(n^{-1})$ in each layer.

\section{Mathematical Theory}
\label{sec:theory}

\begin{figure}[h!]
    \centering
    \includegraphics[scale=0.63]{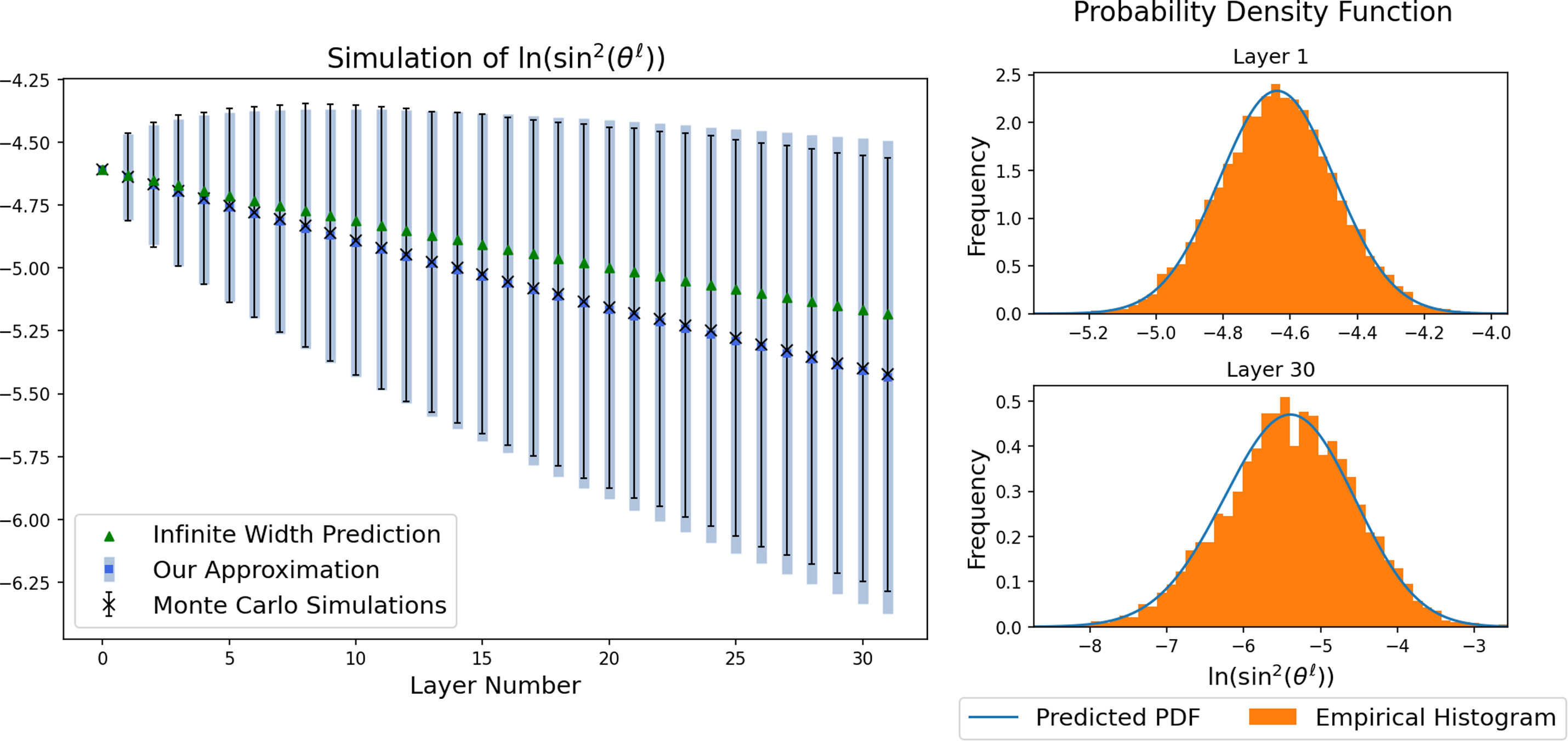}
    \caption{Simulations generated from 5000 independent networks all with uniform hidden layer widths $n_\ell = 256$. Monte Carlo samples are generated by feeding 2 inputs with initial angle $\theta^0=0.1$ into these networks on initialization. Left: We compare the mean and standard deviation of \Cref{approx:Gaussian} vs Monte Carlo samples vs the infinite width prediction as in \Cref{approx:infinite} (which predicts 0 variance and is less accurate at predicting the mean of $\mathbf{E}[\ln(\sin^2(\theta^\ell))]$). Right: Using \Cref{approx:Gaussian}, we compare the predicted probability density function of $\ln(\sin^2(\theta^\ell))$ to the Monte Carlo simulations.}
    \label{updaterule}
\end{figure}

Derivations of \Cref{approx:simple} and \ref{approx:infinite} rely on calculating the joint moments of the ReLU function applied to correlated Gaussian variables (Approximations \ref{approx:simple} and \ref{approx:infinite} are derived in Section 2 and Appendix A.9 of \citet{jakub_nica}, respectively). We provide a very brief overview of the main results of that theory here. A core ingredient in those calculations is the joint moments defined below, which we think of as a family of ``$J$" functions.
\begin{definition}
    Let $G,\; \hat{G}$ be marginally $\mathcal{N}(0,1)$ random variables with correlation $\mathbf{E}[G\hat{G}] = \cos \theta$, and let $\ph(x) = \max \{0,x\}$ be the ReLU activation function. Then, we define an infinite family of $J$ functions as
    $$ J_{a,b}(\theta) = \mathbf{E}[\ph^a(G)\ph^b(\hat{G})].$$
\end{definition}
With this definition of $J_{a,b}(\theta)$, \Cref{approx:infinite} is first derived by the law of large numbers as $\cos(\theta^{\ell+1}) =  2 J_{1,1}(\theta^\ell)$, and then $J_{1,1}$ is explicitly evaluated to obtain \Cref{approx:infinite}.  By using combinatorial expansions, one can get more accurate expansions for $\theta^\ell$ which involve even higher order $J_{a,b}$ (i.e. $a,b\geq 2$) appearing as $\mathcal{O}(n^{-1})$ corrections to the infinite width update rule.  Solving for the higher order, mixed $J$ functions thereby allows us to further correct the infinite width update rule to an update rule that is more accurate for finite width networks. With this approach, we can not only predict the expected value of $\theta^\ell$ at each layer, but we can also study its variance, as shown in \Cref{thm:mean_var_exp} below. 
Consequently, the resulting normal approximation \Cref{approx:Gaussian} matches \emph{both} the mean and \emph{the variance} of actual neural networks remarkably well; see \Cref{updaterule} for Monte Carlo simulations.

\begin{theorem}
\label{thm:mean_var_exp}
Conditionally on the angle $\theta^\ell$ in layer $\ell$, the mean and variance of $\ln \sin^2 (\theta^{\ell+1})$ obey the following limit as the layer width $n_\ell \to \infty$  
\begin{align}
\label{mu_formula2}
\mathbf{E}[\ln \sin^2(\theta^{\ell+1})] =& \mu(\theta^\ell, n_\ell) + \mathcal{O}(n^{-2}_\ell), \quad \var[\ln \sin^2(\theta^{\ell+1})] = \sigma^2(\theta^\ell,n_\ell) + \mathcal{O}({n^{-2}_\ell}),\\ 
\mu(\theta,n) =& \ln\sin^2\theta - \frac{2}{3\pi}\theta - \rho(n) - \frac{8\theta}{15\pi n} -\left(\frac{2}{9\pi^2}-\frac{68}{45\pi^2 n}\right)\theta^2 + \mathcal{O}(\theta^3), \label{eq:mu_asymp_formula} \\
\sigma^2(\theta,n) =&  \frac{8}{n} -  \frac{64}{15\pi}\frac{\theta}{n} - \left(8+\frac{296}{45\pi}\right)\frac{\theta^2}{n} +\mathcal{O}\left( \theta^3\right), \label{eq:si_asymp_formula}
\end{align}
where $\rho(n)$ is as defined in \eqref{eq:rho}.
\end{theorem}



\begin{approximation}
\label{approx:Gaussian}
Conditional on the value of $\theta^\ell$, the angle at layer $\ell+1$ is well approximated by a Gaussian random variable
\begin{equation} \label{eq:update_full}
 \ln \sin^2(\theta^{\ell+1}) \stackrel{d}{\approx} \mathcal{N}(\mu(\theta^\ell,n_\ell), \sigma^2(\theta^\ell,n_\ell)). 
\end{equation}
\end{approximation}

\newpage
\bibliography{sources.bib}

\newpage
\appendix
\section{}

\subsection{Definition of the Network}
\label{app:network_definition}
\begin{table}[h!] 
\begin{center}
\begin{tabular}{ |c|c| } 
\hline
 \textbf{Symbol} & \textbf{Definition} \\ \hline\hline
 $x \in \mathbb{R}^{n_{in}}$ & Input (e.g. training example) in the input dimension $n_{in}\in\mathbb{N}$ \\ \hline
 $\ell \in \mathbb{N}$ & Layer number. $\ell=0$ is the input \\ \hline
 $n_\ell \in \mathbb{N}$ & Width of hidden layer $\ell$ (i.e. number of neurons in layer $\ell$)\\ \hline
 $W^\ell \in \mathbb{R}^{n_{\ell+1} \times n_\ell}$ & Weight matrix for layer $\ell$. Initialized with iid standard Gaussian entries \\
 & $W^{\ell}_{a,b} \sim \mathcal{N}(0,1)$ \\ \hline
 $\varphi: \mathbb{R}^n \to \mathbb{R}^n$ & Entrywise ReLU activation function $\varphi(x)_i = \varphi(x_i)=\max\{x_i,0\} $ \\ \hline
 $z^\ell(x) \in \mathbb{R}^{n_\ell}$ & Pre-activation vector in the $\ell^{\text{th}}$ layer for input $x$ (a.k.a logits of layer $\ell$) \\ 
 & $z^1(x) := W^1x, \quad\quad  z^{\ell+1}(x) := \sqrt{\frac{2}{n_\ell} }W^{\ell+1} \varphi(z^\ell(x)).$\\ \hline
 $\theta^\ell \in [0,\pi]$ & Angle between $\varphi_\alpha^\ell$ and  $\varphi_\beta^\ell$ defined by $\cos(\theta^\ell) := \frac{\langle \varphi_\alpha^\ell, \varphi_\beta^\ell \rangle}{\Vert \varphi_\alpha^\ell \Vert \Vert \varphi_\beta^\ell \Vert }$ \\ \hline
\end{tabular}
\end{center}
\caption{Definition and notation used for fully connected ReLU neural networks.}
\label{tbl:notation}
\end{table}

Given the notation in \Cref{tbl:notation}, a feed-forward ReLU network with $L$ layers is defined as follows:
\begin{equation}
   z^1 = W^1x, \quad\quad  z^{\ell+1} = \sqrt{\frac{2}{n_\ell} }W^{\ell+1} \varphi(z^\ell), \quad\quad f_L(x) = z^L.
\end{equation}

\newpage
\subsection{Network Architectures}
\label{app:network_architectures}
This section details the architectures of the 45 different neural networks used to produce \Cref{fig:simulations}.

\begin{table}[!ht]
    \centering
    \begin{tabular}{|c|c|c|c|c|c|c|c|}
    \hline
        \multirow{2}{*}{\#} & \multirow{2}{*}{Depth} & Avg. & \multicolumn{2}{c|}{\# Parameters}  &  \multicolumn{3}{c|}{Avg. Test Accuracy $\pm$ Standard Deviation}  \\ \cline{4-8}
         &  & Width & (F)MNIST & CIFAR & MNIST & FMNIST & CIFAR \\ \hline \hline
        1 & 2 & 50 & 58880 & 165790 & 0.924 $\pm$ 0.007 & 0.79 $\pm$ 0.02 & 0.211 $\pm$ 0.029 \\ \hline
        2 & 2 & 85 & 57350 & 135510 & 0.837 $\pm$ 0.051 & 0.709 $\pm$ 0.028 & 0.276 $\pm$ 0.011 \\ \hline
        3 & 2 & 200 & 19930 & 54250 & 0.878 $\pm$ 0.009 & 0.721 $\pm$ 0.098 & 0.163 $\pm$ 0.048 \\ \hline
        4 & 2 & 25 & 138300 & 201600 & 0.94 $\pm$ 0.004 & 0.812 $\pm$ 0.009 & 0.229 $\pm$ 0.025 \\ \hline
        5 & 2 & 125 & 31725 & 88925 & 0.89 $\pm$ 0.005 & 0.768 $\pm$ 0.013 & 0.199 $\pm$ 0.027 \\ \hline
        6 & 3 & 25 & 43990 & 114550 & 0.928 $\pm$ 0.008 & 0.812 $\pm$ 0.013 & 0.167 $\pm$ 0.022 \\ \hline
        7 & 3 & 50 & 62830 & 173280 & 0.916 $\pm$ 0.002 & 0.79 $\pm$ 0.012 & 0.224 $\pm$ 0.019 \\ \hline
        8 & 3 & 100 & 59700 & 96756 & 0.952 $\pm$ 0.004 & 0.839 $\pm$ 0.003 & 0.27 $\pm$ 0.016 \\ \hline
        9 & 3 & 67.67 & 87200 & 309900 & 0.924 $\pm$ 0.006 & 0.799 $\pm$ 0.011 & 0.281 $\pm$ 0.011 \\ \hline
        10 & 3 & 50 & 17310 & 189100 & 0.553 $\pm$ 0.181 & 0.599 $\pm$ 0.119 & 0.263 $\pm$ 0.022 \\ \hline
        11 & 4 & 30 & 369400 & 366150 & 0.877 $\pm$ 0.052 & 0.757 $\pm$ 0.026 & 0.192 $\pm$ 0.029 \\ \hline
        12 & 4 & 75 & 99400 & 105060 & 0.957 $\pm$ 0.003 & 0.842 $\pm$ 0.006 & 0.23 $\pm$ 0.025 \\ \hline
        13 & 5 & 21 & 74700 & 51630 & 0.931 $\pm$ 0.005 & 0.811 $\pm$ 0.009 & 0.146 $\pm$ 0.029 \\ \hline
        14 & 6 & 55 & 8840 & 976400 & 0.715 $\pm$ 0.088 & 0.569 $\pm$ 0.146 & 0.337 $\pm$ 0.008 \\ \hline
        15 & 6 & 87.5 & 169400 & 398200 & 0.949 $\pm$ 0.008 & 0.833 $\pm$ 0.007 & 0.332 $\pm$ 0.018 \\ \hline
        16 & 10 & 10 & 79020 & 180010 & 0.951 $\pm$ 0.003 & 0.832 $\pm$ 0.01 & 0.278 $\pm$ 0.018 \\ \hline
        17 & 10 & 100 & 64850 & 122050 & 0.939 $\pm$ 0.004 & 0.824 $\pm$ 0.008 & 0.262 $\pm$ 0.059 \\ \hline
        18 & 10 & 200 & 54170 & 262060 & 0.933 $\pm$ 0.005 & 0.81 $\pm$ 0.014 & 0.335 $\pm$ 0.016 \\ \hline
        19 & 10 & 17.5 & 49920 & 1002300 & 0.794 $\pm$ 0.052 & 0.648 $\pm$ 0.106 & 0.184 $\pm$ 0.026 \\ \hline
        20 & 11 & 34.55 & 518800 & 31720 & 0.955 $\pm$ 0.006 & 0.835 $\pm$ 0.011 & 0.14 $\pm$ 0.037 \\ \hline
        21 & 11 & 35 & 21100 & 269195 & 0.93 $\pm$ 0.005 & 0.823 $\pm$ 0.007 & 0.363 $\pm$ 0.016 \\ \hline
        22 & 13 & 42 & 36420 & 328200 & 0.91 $\pm$ 0.008 & 0.789 $\pm$ 0.01 & 0.364 $\pm$ 0.016 \\ \hline
        23 & 15 & 30 & 41844 & 174100 & 0.92 $\pm$ 0.004 & 0.805 $\pm$ 0.011 & 0.349 $\pm$ 0.015 \\ \hline
        24 & 15 & 50 & 13860 & 235650 & 0.909 $\pm$ 0.005 & 0.8 $\pm$ 0.012 & 0.328 $\pm$ 0.02 \\ \hline
        25 & 15 & 75 & 16580 & 206848 & 0.927 $\pm$ 0.003 & 0.823 $\pm$ 0.007 & 0.359 $\pm$ 0.009 \\ \hline
    \end{tabular}
    \caption{Summary of the architectures of the first 25 neural networks used in \Cref{fig:simulations}, as well as their performance on the test datasets. Note that the number of parameters differs between the (F)MNIST and CIFAR-10 datasets due to the fact that CIFAR-10 images are in colour requiring 3 colour channels, while the MNIST and FMNIST images are in grayscale. This table is continued in \Cref{tab:summary_2}. }
    \label{tab:summary_1}
\end{table}

\begin{table}[!ht]
    \centering
    \begin{tabular}{|c|c|c|c|c|c|c|c|}
    \hline
        \multirow{2}{*}{\#} & \multirow{2}{*}{Depth} & Avg. & \multicolumn{2}{c|}{\# Parameters}  &  \multicolumn{3}{c|}{Average Score $\pm$ Standard Deviation}  \\ \cline{4-8}
         &  & Width & (F)MNIST & CIFAR & MNIST & FMNIST & CIFAR \\ \hline \hline
        26 & 16 & 35 & 42200 & 159100 & 0.943 $\pm$ 0.004 & 0.838 $\pm$ 0.004 & 0.343 $\pm$ 0.021 \\ \hline
        27 & 16 & 22.5 & 198800 & 656400 & 0.963 $\pm$ 0.003 & 0.845 $\pm$ 0.01 & 0.37 $\pm$ 0.016 \\ \hline
        28 & 20 & 25 & 94900 & 323700 & 0.955 $\pm$ 0.002 & 0.843 $\pm$ 0.006 & 0.367 $\pm$ 0.006 \\ \hline
        29 & 20 & 50 & 60416 & 62340 & 0.951 $\pm$ 0.003 & 0.837 $\pm$ 0.005 & 0.163 $\pm$ 0.058 \\ \hline
        30 & 20 & 37.5 & 44700 & 156600 & 0.948 $\pm$ 0.003 & 0.834 $\pm$ 0.008 & 0.346 $\pm$ 0.028 \\ \hline
        31 & 23 & 31.30 & 194550 & 598200 & 0.927 $\pm$ 0.005 & 0.788 $\pm$ 0.008 & 0.17 $\pm$ 0.004 \\ \hline
        32 & 25 & 15 & 64050 & 48180 & 0.951 $\pm$ 0.002 & 0.84 $\pm$ 0.004 & 0.186 $\pm$ 0.071 \\ \hline
        33 & 25 & 75 & 55160 & 125880 & 0.899 $\pm$ 0.014 & 0.748 $\pm$ 0.033 & 0.274 $\pm$ 0.048 \\ \hline
        34 & 25 & 150 & 53760 & 64390 & 0.782 $\pm$ 0.077 & 0.676 $\pm$ 0.064 & 0.206 $\pm$ 0.041 \\ \hline
        35 & 28 & 35.71 & 74715 & 78300 & 0.953 $\pm$ 0.001 & 0.844 $\pm$ 0.001 & 0.244 $\pm$ 0.075 \\ \hline
        36 & 30 & 15 & 60860 & 152380 & 0.819 $\pm$ 0.08 & 0.719 $\pm$ 0.033 & 0.17 $\pm$ 0.02 \\ \hline
        37 & 30 & 30 & 18630 & 145280 & 0.862 $\pm$ 0.08 & 0.772 $\pm$ 0.017 & 0.168 $\pm$ 0.02 \\ \hline
        38 & 30 & 100 & 34360 & 146680 & 0.941 $\pm$ 0.003 & 0.826 $\pm$ 0.009 & 0.165 $\pm$ 0.022 \\ \hline
        39 & 30 & 26.67 & 659100 & 118560 & 0.932 $\pm$ 0.014 & 0.785 $\pm$ 0.011 & 0.175 $\pm$ 0.007 \\ \hline
        40 & 30 & 31.67 & 18435 & 52755 & 0.313 $\pm$ 0.131 & 0.349 $\pm$ 0.109 & 0.158 $\pm$ 0.026 \\ \hline
        41 & 35 & 40 & 86160 & 276600 & 0.753 $\pm$ 0.074 & 0.586 $\pm$ 0.11 & 0.148 $\pm$ 0.029 \\ \hline
        42 & 35 & 75 & 250800 & 450525 & 0.725 $\pm$ 0.163 & 0.608 $\pm$ 0.077 & 0.165 $\pm$ 0.007 \\ \hline
        43 & 40 & 50 & 137200 & 251600 & 0.522 $\pm$ 0.141 & 0.513 $\pm$ 0.089 & 0.167 $\pm$ 0.007 \\ \hline
        44 & 40 & 75 & 278925 & 422400 & 0.467 $\pm$ 0.123 & 0.466 $\pm$ 0.09 & 0.161 $\pm$ 0.022 \\ \hline
        45 & 50 & 50 & 162200 & 177680 & 0.242 $\pm$ 0.064 & 0.22 $\pm$ 0.042 & 0.161 $\pm$ 0.019 \\ \hline
    \end{tabular}
    \caption{Continuation of \Cref{tab:summary_1} for networks 26 through 45.}
    \label{tab:summary_2}
\end{table}

\begin{table}[!ht]
    \centering
    \begin{tabular}{|c|l|}
    \hline
        \# & Hidden Layer Widths \\ \hline
        1 & 50, 50 \\ \hline
        2 & 85, 85 \\ \hline
        3 & 200, 200 \\ \hline
        4 & 20, 30 \\ \hline
        5 & 100, 150 \\ \hline
        6 & 25, 25, 25 \\ \hline
        7 & 50, 50, 50 \\ \hline
        8 & 100, 100, 100 \\ \hline
        9 & 64, 75, 64 \\ \hline
        10 & 75, 50, 25 \\ \hline
        11 & 40, 40, 20, 20 \\ \hline
        12 & 50, 100, 100, 50 \\ \hline
        13 & 15, 15, 15, 30, 30 \\ \hline
        14 & 80, 70, 60, 50, 40, 30 \\ \hline
        15 & 25, 50, 75, 100, 125, 150 \\ \hline
        16 & 10, 10, 10, 10, 10, 10, 10, 10, 10, 10 \\ \hline
        17 & 100, 100, 100, 100, 100, 100, 100, 100, 100, 100 \\ \hline
        18 & 200, 200, 200, 200, 200, 200, 200, 200, 200, 200 \\ \hline
        19 & 20, 20, 20, 20, 20, 15, 15, 15, 15, 15 \\ \hline
        20 & 55, 30, 30, 30, 30, 30, 30, 30, 30, 30, 55 \\ \hline
        21 & 40, 39, 38, 37, 36, 35, 34, 33, 32, 31, 30 \\ \hline
        22 & 24, 27, 30, 33, 36, 39, 42, 45, 48, 51, 54, 57, 60 \\ \hline
        23 & 30, 30, 30, 30, 30, 30, 30, 30, 30, 30, 30, 30, 30, 30, 30 \\ \hline
        24 & 50, 50, 50, 50, 50, 50, 50, 50, 50, 50, 50, 50, 50, 50, 50 \\ \hline
        25 & 75, 75, 75, 75, 75, 75, 75, 75, 75, 75, 75, 75, 75, 75, 75 \\ \hline
    \end{tabular}
    \caption{Ordered list of hidden layer widths for the first 25 networks used in \Cref{fig:simulations}. This table is continued in \Cref{tab:widths_2}.}
    \label{tab:widths_1}
\end{table}

\begin{table}[!ht]
    \centering
    \begin{tabular}{|c|p{15cm}|}
    \hline
        \# & Hidden Layer Widths \\ \hline
        26 & 50, 48, 46, 44, 42, 40, 38, 36, 34, 32, 30, 28, 26, 24, 22, 20 \\ \hline
        27 & 15, 16, 17, 18, 19, 20, 21, 22, 23, 24, 25, 26, 27, 28, 29, 30 \\ \hline
        28 & 25, 25, 25, 25, 25, 25, 25, 25, 25, 25, 25, 25, 25, 25, 25, 25, 25, 25, 25, 25 \\ \hline
        29 & 50, 50, 50, 50, 50, 50, 50, 50, 50, 50, 50, 50, 50, 50, 50, 50, 50, 50, 50, 50 \\ \hline
        30 & 45, 45, 45, 45, 45, 40, 40, 40, 40, 40, 35, 35, 35, 35, 35, 30, 30, 30, 30, 30 \\ \hline
        31 & 40, 40, 40, 40, 40, 40, 40, 40, 40, 40, 40, 40, 40, 20, 20, 20, 20, 20, 20, 20, 20, 20, 20 \\ \hline
        32 & 15, 15, 15, 15, 15, 15, 15, 15, 15, 15, 15, 15, 15, 15, 15, 15, 15, 15, 15, 15, 15, 15, 15, 15, 15 \\ \hline
        33 & 75, 75, 75, 75, 75, 75, 75, 75, 75, 75, 75, 75, 75, 75, 75, 75, 75, 75, 75, 75, 75, 75, 75, 75, 75 \\ \hline
        34 & 150, 150, 150, 150, 150, 150, 150, 150, 150, 150, 150, 150, 150, 150, 150, 150, 150, 150, 150, 150, 150, 150, 150, 150, 150 \\ \hline
        35 & 25, 25, 25, 25, 50, 50, 50, 50, 25, 25, 25, 25, 50, 50, 50, 50, 25, 25, 25, 25, 50, 50, 50, 50, 25, 25, 25, 25 \\ \hline
        36 & 15, 15, 15, 15, 15, 15, 15, 15, 15, 15, 15, 15, 15, 15, 15, 15, 15, 15, 15, 15, 15, 15, 15, 15, 15, 15, 15, 15, 15, 15 \\ \hline
        37 & 30, 30, 30, 30, 30, 30, 30, 30, 30, 30, 30, 30, 30, 30, 30, 30, 30, 30, 30, 30, 30, 30, 30, 30, 30, 30, 30, 30, 30, 30 \\ \hline
        38 & 100, 100, 100, 100, 100, 100, 100, 100, 100, 100, 100, 100, 100, 100, 100, 100, 100, 100, 100, 100, 100, 100, 100, 100, 100, 100, 100, 100, 100, 100 \\ \hline
        39 & 40, 40, 40, 40, 40, 20, 20, 20, 20, 20, 20, 20, 20, 20, 20, 20, 20, 20, 20, 20, 20, 20, 20, 20, 20, 40, 40, 40, 40, 40 \\ \hline
        40 & 40, 40, 40, 40, 40, 30, 30, 30, 30, 30, 30, 30, 30, 30, 30, 30, 30, 30, 30, 30, 30, 30, 30, 30, 30, 30, 30, 30, 30, 30 \\ \hline
        41 & 40, 40, 40, 40, 40, 40, 40, 40, 40, 40, 40, 40, 40, 40, 40, 40, 40, 40, 40, 40, 40, 40, 40, 40, 40, 40, 40, 40, 40, 40, 40, 40, 40, 40, 40 \\ \hline
        42 & 75, 75, 75, 75, 75, 75, 75, 75, 75, 75, 75, 75, 75, 75, 75, 75, 75, 75, 75, 75, 75, 75, 75, 75, 75, 75, 75, 75, 75, 75, 75, 75, 75, 75, 75 \\ \hline
        43 & 50, 50, 50, 50, 50, 50, 50, 50, 50, 50, 50, 50, 50, 50, 50, 50, 50, 50, 50, 50, 50, 50, 50, 50, 50, 50, 50, 50, 50, 50, 50, 50, 50, 50, 50, 50, 50, 50, 50, 50 \\ \hline
        44 & 75, 75, 75, 75, 75, 75, 75, 75, 75, 75, 75, 75, 75, 75, 75, 75, 75, 75, 75, 75, 75, 75, 75, 75, 75, 75, 75, 75, 75, 75, 75, 75, 75, 75, 75, 75, 75, 75, 75, 75 \\ \hline
        45 & 50, 50, 50, 50, 50, 50, 50, 50, 50, 50, 50, 50, 50, 50, 50, 50, 50, 50, 50, 50, 50, 50, 50, 50, 50, 50, 50, 50, 50, 50, 50, 50, 50, 50, 50, 50, 50, 50, 50, 50, 50, 50, 50, 50, 50, 50, 50, 50, 50, 50 \\ \hline
    \end{tabular}
    \caption{Continuation of \Cref{tab:widths_1} for networks 26 through 45.}
    \label{tab:widths_2}
\end{table}

\end{document}